\pdfoutput=1

\documentclass[11pt]{article}

\usepackage{acl}

\usepackage{times}
\usepackage{latexsym}

\usepackage[T1]{fontenc}

\usepackage[utf8]{inputenc}
\usepackage{microtype}

\title{DEGREE: A Data-Efficient Generation-Based Event Extraction Model}

\author{I-Hung Hsu\thanks{\; The authors contribute equally.}$^{\;\;\dagger}$ \ \ \ Kuan-Hao Huang\footnotemark[1]$^{\;\;\ddagger}$ \ \ \ Elizabeth Boschee$^{\dagger}$ \ \ \ Scott Miller$^{\dagger}$ \\
{\bf Premkumar Natarajan$^{\dagger}$ \ \ \ Kai-Wei Chang$^{\ddagger}$ \ \ \ Nanyun Peng$^{\dagger \ddagger}$}\\
$^{\dagger}$Information Science Institute, University of Southern California \\
$^{\ddagger}$Computer Science Department, University of California, Los Angeles \\ 
\texttt{\{ihunghsu, boschee, smiller, pnataraj\}@isi.edu} \\
\texttt{\{khhuang, kwchang, violetpeng\}@cs.ucla.edu} \\}
  
\usepackage[flushleft]{threeparttable}
\usepackage{todonotes}
\usepackage{enumitem}
\usepackage{booktabs}
\usepackage{subcaption}
\usepackage[nointegrals]{wasysym}
\usepackage{tikz}
\usepackage{tabularx}
\usepackage{tabulary}
\usepackage{listings}
\usepackage{color}
\usepackage{multirow}
\usepackage{colortbl}
\usepackage{framed}
\usepackage{pifont}
\usepackage{dsfont}
\usepackage{xcolor}
\usepackage{amsmath,amsthm}
\usepackage{amssymb}
\usepackage{graphicx}
\usepackage{textcomp}
\usepackage{tcolorbox}
\usepackage{soul}
\usepackage{lipsum}
\usepackage{makecell}
\usepackage{amsfonts}

\definecolor{DarkRed}{RGB}{130,25,0}

\newcommand{\ignore}[1]{}

\definecolor{mypurple}{RGB}{121,55,196}
\definecolor{myblue}{RGB}{60,177,245}
\definecolor{myorange}{RGB}{243,206,84}
\definecolor{mygreen}{RGB}{41,222,35}
\definecolor{light-gray}{gray}{0.9}

\newcommand{\Skip}[1]{{}}
\newcommand{\model}[1]{\textsc{Degree}}
\newcommand{\edmodel}[1]{\textsc{Degree(ED)}}
\newcommand{\eaemodel}[1]{\textsc{Degree(EAE)}}
\newcommand{\pipemodel}[1]{\textsc{Degree(Pipe)}}
\newcommand{\eventdef}{$\mathsf{event\ type\ definition}$}
\newcommand{\Eventdef}{$\mathsf{Event\ type\ definition}$}
\newcommand{\eventkey}{$\mathsf{event\ keywords}$}
\newcommand{\Eventkey}{$\mathsf{Event\ keywords}$}
\newcommand{\edtemp}{$\mathsf{ED\ template}$}

\newcommand{\querytrig}{$\mathsf{query\ trigger}$}
\newcommand{\Querytrig}{$\mathsf{Query\ trigger}$}
\newcommand{\eaetemp}{$\mathsf{EAE\ template}$}
\newcommand{\eaetemps}{$\mathsf{EAE\ templates}$}
\newcommand{\etoetemp}{$\mathsf{E2E\ template}$}
\newcommand{\etoetemps}{$\mathsf{E2E\ templates}$}
\setuldepth{Berlin}
\newcolumntype{x}[1]{>{\arraybackslash\hspace{0pt}}m{#1}}
\newcolumntype{y}[1]{>{\centering\arraybackslash\hspace{0pt}}m{#1}}

\begin{document}
\maketitle
\begin{abstract}
Event extraction requires high-quality expert human annotations, which are usually expensive. Therefore, learning a \emph{data-efficient} event extraction model that can be trained with \emph{only a few} labeled examples has become a crucial challenge. In this paper, we focus on \emph{low-resource end-to-end} event extraction and propose \model{}, a data-efficient model that formulates event extraction as a conditional generation problem. Given a passage and a manually designed prompt, \model{} learns to summarize the events mentioned in the passage into a natural sentence that follows a predefined pattern. The final event predictions are then extracted from the generated sentence with a deterministic algorithm. \model{} has three advantages to learn well with less training data. First, our designed prompts provide semantic guidance for \model{} to leverage \emph{label semantics} and thus better capture the event arguments. Moreover, \model{} is capable of using additional \emph{weakly-supervised} information, such as the description of events encoded in the prompts. Finally, \model{} learns triggers and arguments jointly in an \emph{end-to-end manner}, which encourages the model to better utilize the shared knowledge and dependencies among them. Our experimental results demonstrate the strong performance of \model{} for low-resource event extraction.
\end{abstract}
\section{Introduction}
\label{sec:intro}
\begin{figure}[t!]
    \centering
    \includegraphics[trim=0cm 0cm 0cm 0cm, clip, width=1.0\columnwidth]{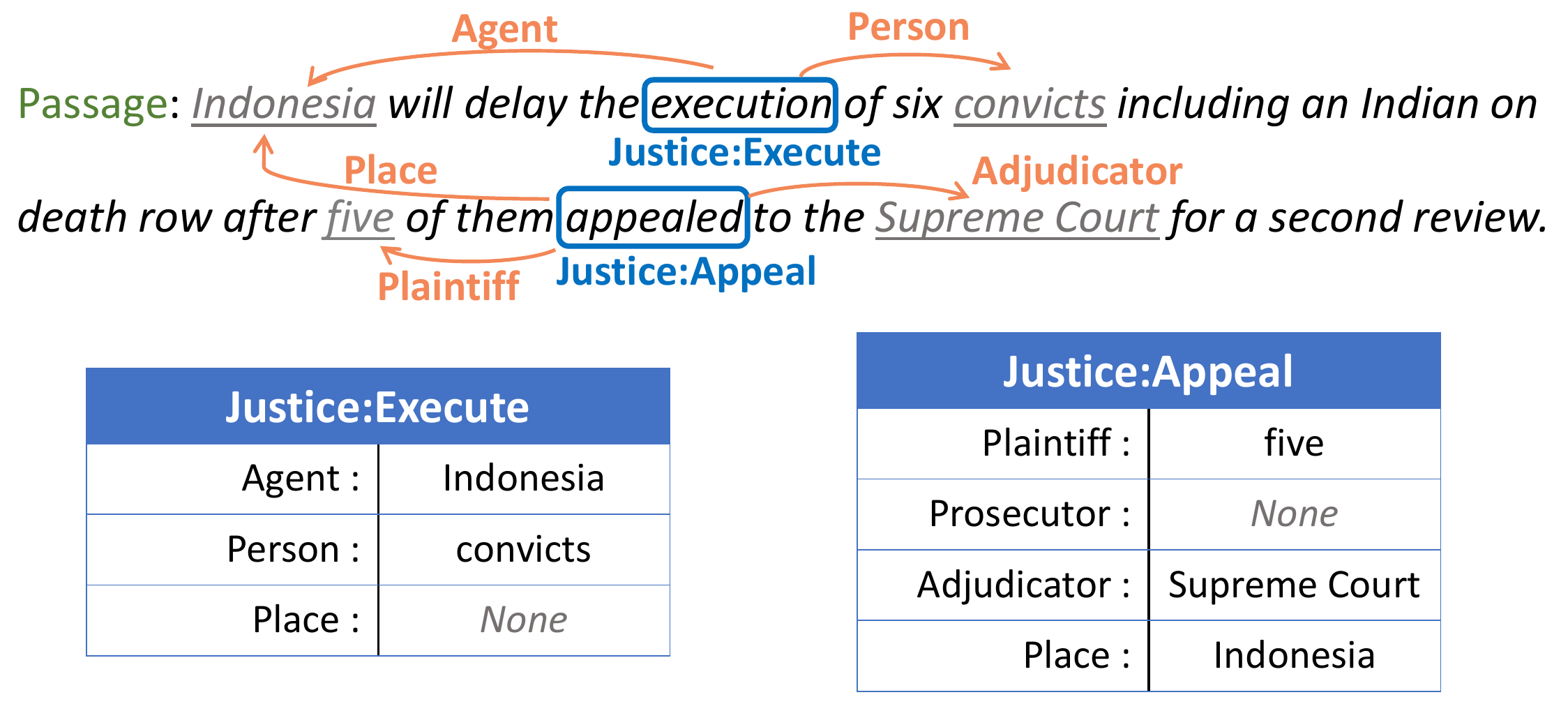}
    \caption{Two examples of events (\emph{Justice:Execute} and \emph{Justice:Appeal}) extracted from the given passage.
    }
    \label{fig:intro_semantic}
    \vspace{-1.2em}
\end{figure}

Event extraction (EE) aims to extract events, each of which consists of a trigger and several participants (arguments) with their specific roles, from a given passage. 
For example, in Figure~\ref{fig:intro_semantic}, a \textit{Justice:Execute} event is triggered by the word \textit{``execution''} and this event contains three argument roles, including an \textit{Agent} (\textit{Indonesia}) who carries out the execution, a \textit{Person} who is executed (\textit{convicts}), and a \textit{Place} where the event occurs (not mentioned in the passage). 
Previous work \cite{DBLP:conf/acl/YangFQKL19, DBLP:journals/corr/abs-2109-12383} usually divides EE into two subtasks: 
(1) \textbf{event detection}, which identifies event triggers and their types, and (2) \textbf{event argument extraction}, which extracts the arguments and their roles for given event triggers.
EE has been shown to benefit a wide range of applications, e.g., building knowledge graphs~\cite{DBLP:conf/www/ZhangLPSL20}, question answering ~\cite{berant-etal-2014-modeling, DBLP:conf/emnlp/HanHSBNRP21}, and other downstream studies~\cite{DBLP:conf/conll/HanHYGWP19, hogenboom2016survey, DBLP:conf/acl/SunP20}.

\begin{figure*}[t!]
    \centering
    \includegraphics[trim=0cm 0cm 0cm 0cm, clip, width=0.9\textwidth]{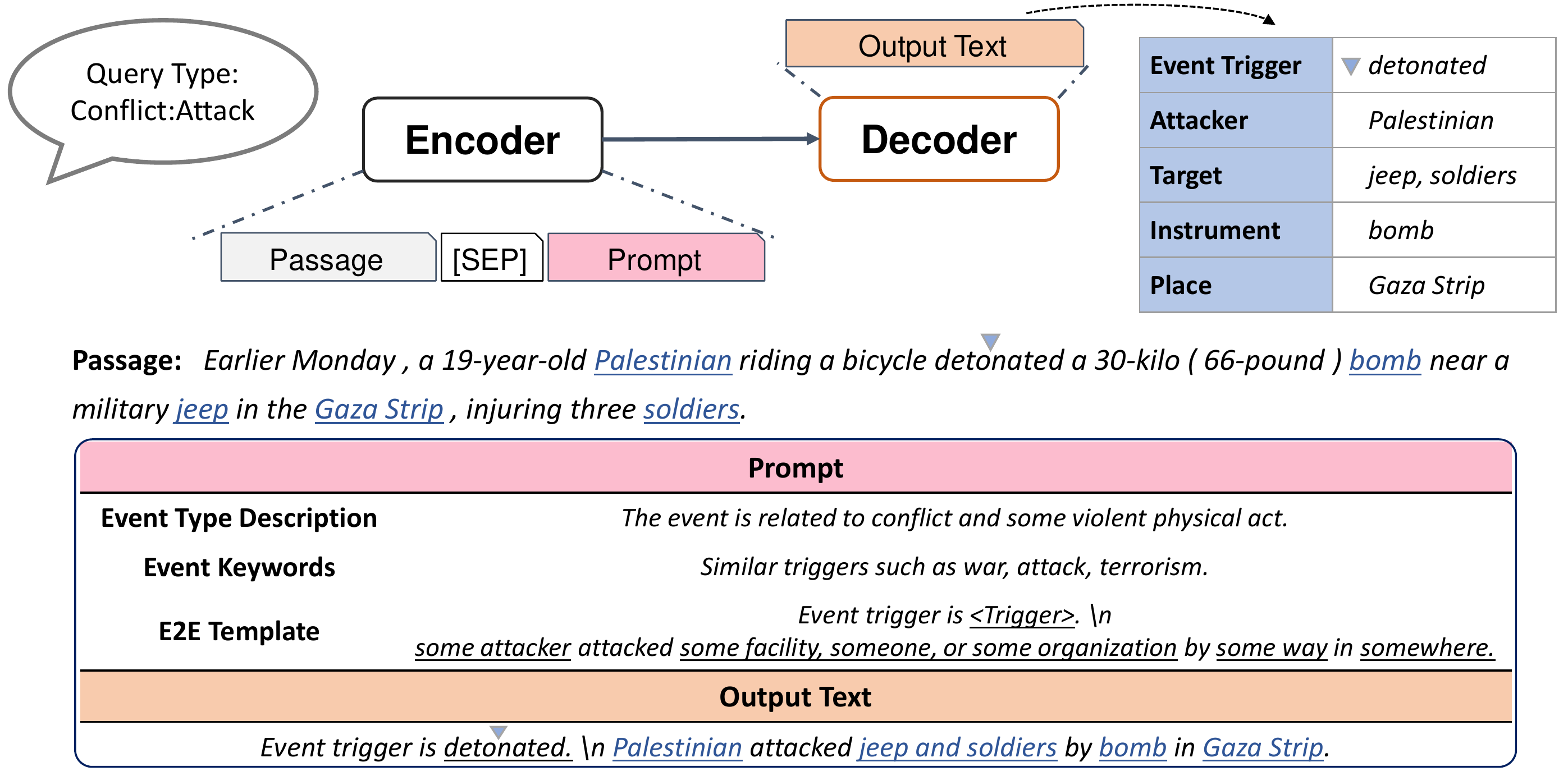}
    \caption{An illustration of \model{} for predicting a \textit{Contact:Attack} event. 
    The input of \model{} consists of the given passage and our design prompt that contains an event type description, event keywords, and a \etoetemp{}.
    \model{} is trained to generate an output to fill in the placeholders (underlined words) in the \etoetemp{} with triggers and arguments.
    The final event prediction is then decoded from the generated output.}
    \label{fig:mthd_overview}
    \vspace{-0.8em}
\end{figure*}

Most prior works on EE rely on a large amount of annotated data for training \cite{nguyen-grishman-2015-event, Nguyen16jrnn, han2019joint, Du20qa, huang2020event, huang2021document, Paolini21tacl}.
However, high-quality event annotations are expensive to obtain.
For example, the ACE 2005 corpus~\cite{doddington-etal-2004-automatic}, one of the most widely used EE datasets, requires two rounds of annotations by linguistics experts. 
The high annotation costs make these models hard to be extended to new domains and new event types.
Therefore, how to learn a \emph{data-efficient} EE model trained with \emph{only a few} annotated examples is a crucial challenge.

In this paper, we focus on \emph{low-resource} event extraction, where only a small amount of training examples are available for training.
We propose \model{} (\textbf{D}ata-\textbf{E}fficient \textbf{G}ene\textbf{R}ation-Based \textbf{E}vent \textbf{E}xtraction), a generation-based model that takes a passage and a manually designed prompt as the input, and learns to summarize the passage into a natural sentence following a predefined template, as illustrated in Figure~\ref{fig:mthd_overview}.
The event triggers and arguments can then be extracted from the generated sentence by using a deterministic algorithm. 

\model{} enjoys the following advantages to learn well with less training data.
First, 
the framework provides \emph{label semantics} via the designed template in the prompts.
As the example in Figure~\ref{fig:mthd_overview} shows, the word \emph{``somewhere''} in the prompt guides the model to predict words being similar to \emph{location} for the role \emph{Place}.
In addition, the sentence structure of the template and the word \emph{``attacked''} depict the semantic relation between the role \emph{Attacker} and the role \emph{Target}.
With these kinds of guidance, \model{} can make more accurate predictions with less training examples.
Second, the prompts can incorporate additional weak-supervision signal about the task, such as the description of the event and similar keywords. These resources are usually readily available. For example, in our experiments, we take the information from the annotation guideline, which is provided along with the dataset.
This information facilitates \model{} to learn under a low-resource situation.
Finally, \model{} is designed for end-to-end event extraction and can solve event detection and event argument extraction at the same time. Leveraging the shared knowledge and dependencies between the two tasks makes our model more data-efficient.

Existing works on EE usually have only one or two of above-mentioned advantages.
For example, previous classification-based models \cite{Nguyen16jrnn, wang-etal-2019-hmeae, yang-etal-2019-exploring, Wadden19dygiepp, Lin20oneie} can hardly encode label semantics and other weak supervision signals.
Recently proposed generation-based models for event extraction solved the problem in a pipeline fashion; therefore, they cannot leverage shared knowledge between subtasks \cite{Paolini21tacl,li2021document}.
Furthermore, their generated outputs are not natural sentences, which hinders the utilization of label semantics \cite{Paolini21tacl, text2event}.
As a result, our model \model{} can achieve significantly better performance than prior approaches on low-resource event extraction, as we will demonstrate in Section~\ref{sec:low_data_result}. 

Our contributions can be summarized as follows:
\begin{itemize}[topsep=1pt, itemsep=-2pt, leftmargin=13pt]
    \item We propose \model{}, a generation-based event extraction model that learns well with less data by better incorporating label semantics and shared knowledge between subtasks (Section~\ref{sec:method}).
    
    \item Experiments on ACE 2005 \cite{doddington-etal-2004-automatic} and ERE-EN \cite{DBLP:conf/aclevents/SongBSRMEWKRM15} demonstrate the strong performance of \model{} in the low-resource setting (Section~\ref{sec:low_data_result}).
    
    \item We present comprehensive ablation studies in both the low-resource and the high-resource setting to better understand the strengths and weaknesses of our model (Section~\ref{sec:ablation}).

\end{itemize}

Our code and models can be found at \url{https://github.com/PlusLabNLP/DEGREE}.


\section{Data-Efficient Event Extraction}
\label{sec:method}

We introduce \model{}, a generation-based model for low-resource event extraction.
Unlike previous works \cite{DBLP:conf/acl/YangFQKL19,li2021document}, which separate event extraction into two pipelined tasks (event detection and event argument extraction), \model{} is designed for the end-to-end event extraction and predict event triggers and arguments at the same time.

\subsection{The \model{} Model}

We formulate event extraction as a conditional generation problem.
As illustrated in Figure~\ref{fig:mthd_overview}, given a passage and our designed prompt, \model{} generates an output following a particular format. The final predictions of event triggers and argument roles can be then parsed from the generated output with a deterministic algorithm.
Compared to previous classification-based models \cite{wang-etal-2019-hmeae, yang-etal-2019-exploring, Wadden19dygiepp,Lin20oneie}, the generation framework provides a flexible way to include additional information and guidance.
By designing appropriate prompts, we encourage \model{} to better capture the dependencies between entities and, therefore, to reduce the number of training examples needed.

The desired prompt not only provides information but also defines the output format.
As shown in Figure~\ref{fig:mthd_overview}, it contains the following components:
\begin{itemize}[topsep=3pt, itemsep=-2pt, leftmargin=13pt]
    \item $\boldsymbol{\mathsf{Event\ type\ definition}}$ describes the definition for the given event type.\footnote{The definition can be derived from the annotation guidelines, which are provided along with the datasets.} For example, \emph{``The event is related to conflict and some violent physical act.''} describes a \emph{Conflict:Attack} event.
    
    \item $\boldsymbol{\mathsf{Event\ keywords}}$ presents some words that are semantically related to the given event type. For example, \emph{war}, \emph{attack}, and \emph{terrorism} are three event keywords for the \emph{Conflict:Attack} event. In practice, we collect three words that appear as the triggers in the example sentences from the annotation guidelines.
    
    \item $\boldsymbol{\mathsf{E2E\ template}}$ defines the expected output format and can be separated into two parts. 
    The first part is called \edtemp{}, which is designed as
    \textit{``Event trigger is $<$Trigger$>$''}, where \textit{``$<$Trigger$>$''} is a special token serving as a placeholder.
    The second part is the \eaetemp{}, which differs based on the given event type.
    For example, in Figure~\ref{fig:mthd_overview}, the \eaetemp{} for a \textit{Conflict:Attack} event is \textit{``\ul{some attacker} attacked \ul{some facility, someone, or some organization} by \ul{some way} in \ul{somewhere}''}. Each underlined string starting with \textit{``some-''} serves as a placeholder corresponding to an argument role for a \textit{Conflict:Attack} event. For instance, \textit{``some way''} corresponds to the role \textit{Instrument} and \textit{``somewhere''} corresponds to the role \textit{Place}.
    Notice that every event type has its own \eaetemp{}. We list three \eaetemps{} in Table~\ref{tab:example-temp-ace}. The full list of \eaetemps{} and the construction details can be found in Appendix~\ref{sec:eaetemp_list}.
\end{itemize}

\begin{table*}[t]
\centering
\small
\setlength{\tabcolsep}{5pt}
\begin{tabular}{|x{.3\textwidth}|x{.65\textwidth}|}
    \hline
    \textbf{Event Type} & \textbf{EAE Template} \\
    \hline
Life:Divorce & somebody divorced in somewhere. \\
\hline
Transaction:Transfer-Ownership & someone got something from some seller in somewhere. \\
\hline
Justice:Sue & somebody was sued by some other in somewhere. The adjudication was judged by some adjudicator. \\
\hline
\end{tabular}
\caption{
Three examples of \eaetemps{} for the ACE 2005 corpus.}
\label{tab:example-temp-ace}
\end{table*}

\subsection{Training}
The training objective of \model{} is to generate an output that replaces the placeholders in \etoetemp{} with the gold labels. 
Take Figure~\ref{fig:mthd_overview} as an example, \model{} is expected to replace \textit{``$<$Trigger$>$''} with the gold trigger (\emph{detonated}), replace \textit{``some attacker''} with the gold argument for role \emph{Attacker} (\emph{Palestinian}), and replace \textit{``some way''} with the gold argument for role \emph{Instrument} (\emph{bomb}).
If there are multiple arguments for the same role, they are concatenated with \textit{``and''}; if there is no predicted argument for one role, the model should keep the corresponding placeholder (i.e, \textit{``some-''} in the \etoetemp{}).
For the case that there are multiple triggers for the given event type in the input passage, \model{} is trained to generate the output text that contains multiple \etoetemp{} such that each \etoetemp{} corresponds to one trigger and its argument roles.
The hyperparameter settings are detailed in Appendix~\ref{sec:training}.

\subsection{Inference}
We enumerate all event types and generate an output for each event type.
After we obtain the generated sentences, we compare the outputs with \etoetemp{} to determine the predicted triggers and arguments in string format.
Finally, we apply string matching to convert the predicted string to span offsets in the passage.
If the predicted string appears in the passage multiple times, we choose all span offsets that match for trigger predictions and choose the one closest to the given trigger span for argument predictions.

\subsection{Discussion}
Notice that the \etoetemp{} plays an important role for \model{}.
First, it serves as the control signal and defines the expected output format.
Second, it provides label semantics to help \model{} make accurate predictions.
Those placeholders (words starting with \textit{``some-''}) in the \etoetemp{} give \model{} some hints about the entity types of arguments.
For instance, when seeing \emph{``somewhere''}, \model{} tends to generate a location rather than a person. In addition, the words other than \textit{``some-''} describe the relationships between roles. 
For example, \model{} knows the relationship between the role \emph{Attacker} and the role \emph{Target} (who is attacking and who is attacked) due to \etoetemp{}.
This guidance helps \model{} learn the dependencies between entities.

Unlike previous generation-based approaches \cite{Paolini21tacl,li-etal-2020-event, DBLP:conf/emnlp/HuangTP21}, we intentionally write \etoetemps{} in natural sentences.
This not only uses label semantics better but also makes the model easier to leverage the knowledge from the pre-trained decoder. 
In Section~\ref{sec:ablation}, we will provide experiments to demonstrate the advantage of using natural sentences.


\paragraph{Cost of template constructing.}
\model{} does require human effort to design the templates; however, writing those templates is much easier and more effortless than collecting complicated event annotations.
As shown in Table~\ref{tab:example-temp-ace}, we keep the \eaetemps{} as simple and short as possible.
Therefore, it takes only about one minute for people who are not linguistic experts to compose a template.
In fact, several prior works \cite{Liu2020rceeer,Du20qa,li-etal-2020-event} also use constructed templates as weakly-supervised signals to improve models.
In Section~\ref{sec:ablation}, we will study how different templates affect the performance.

\paragraph{Efficiency Considerations.}
\model{} requires to enumerate all event types during inference, which could cause efficiency considerations when extending to applications that contain many event types.
This issue is minor for our experiments on the two datasets (ACE 2005 and ERE-EN), which are relatively small scales in terms of the number of event types. 
Due to the high cost of annotations, there is hardly any public datasets for end-to-end event extraction on a large scale,\footnote{To the best of our knowledge, MAVEN~\cite{DBLP:conf/emnlp/WangWHJHLLLLZ20} is the only publicly available large-scale event dataset. However, the dataset only focuses on event detection without considering event arguments.} and we cannot provide a more thorough studies when the experiments scale up.
We leave the work on benchmarking and improving the efficiency of \model{} in the scenario considering more diverse and comprehensive types of events as future work. 

\subsection{\model{} in Pipeline Framework}
\model{} is flexible and can be easily modified to \pipemodel{}, which first focuses event detection~(ED) and then solves event argument extraction~(EAE).
\pipemodel{} consists of two models: (1) \edmodel{}, which aims to exact event triggers for the given event type, and (2) \eaemodel{}, which identifies argument roles for the given event type and the corresponding trigger.
\edmodel{} and \eaemodel{} are similar to \model{} but with different prompts and output formats.
We describe the difference as follows.

\paragraph{\edmodel{}.}
\label{sec:ed}
The prompt of \edmodel{} contains the following components:

\begin{itemize}[topsep=3pt, itemsep=-2pt, leftmargin=13pt]
    \item $\boldsymbol{\mathsf{Event\ type\ definition}}$ is the same as the ones for \model{}.
    \item $\boldsymbol{\mathsf{Event\ keywords}}$ is the same as the one for \model{}.
    \item $\boldsymbol{\mathsf{ED\ template}}$ is  designed as  \textit{``Event trigger is $<$Trigger$>$''}, which is actually the first part of the \etoetemp{}.
\end{itemize}

\noindent Similar to \model{}, the objective of \edmodel{} is to generate an output that replaces \textit{``$<$Trigger$>$''} in the \edtemp{} with event triggers.

\paragraph{\eaemodel{}.}
\label{sec:eae}
The prompt of \eaemodel{} contains the following components:

\begin{itemize}[topsep=3pt, itemsep=-2pt, leftmargin=13pt]
    \item $\boldsymbol{\mathsf{Event\ type\ definition}}$ is the same as the one for \model{}.
    
    \item $\boldsymbol{\mathsf{Query\ trigger}}$ is a string that indicates the trigger word for the given event type.
    For example, \emph{``The event trigger word is detonated''} points out that \emph{``detonated''} is the given trigger.
    
    \item $\boldsymbol{\mathsf{EAE\ template}}$ is an event-type-specific template mentioned previously. 
    It is actually the second part of \etoetemp{}.

\end{itemize}

\noindent Similar to \model{}, the goal for \eaemodel{} is to generate an outputs that replace the placeholders in \eaetemp{} with event arguments.

In Section~\ref{sec:low_data_result}, we will compare \model{} with \pipemodel{} to study the benefit of dealing with event extraction in an end-to-end manner under the low-resource setting.

\begin{table*}[t!]
\centering
\small
\resizebox{1.0\textwidth}{!}{
\setlength{\tabcolsep}{2.5pt}
\begin{tabular}{|l|c|cccccc|cccccc|cccccc|}
    \hline
    \multicolumn{20}{|c|}{\textbf{Trigger Classification F1-Score (\%)}} \\
    \hline
    \multirow{2}{*}{Model} & \multirow{2}{*}{Type} & \multicolumn{6}{c|}{ACE05-E} & \multicolumn{6}{c|}{ACE05-E$^{+}$} & \multicolumn{6}{c|}{ERE-EN} \\
    \cline{3-20}
    & & 1\% & 3\% & 5\% & 10\% & 20\% & 30\% & 1\% & 3\% & 5\% & 10\% & 20\% & 30\% & 1\% & 3\% & 5\% & 10\% & 20\% & 30\% \\
    \hline
    BERT\_QA & Cls
    & 20.5 & 40.2 & 42.5 & 50.1 & 61.5 & 61.3 & -  & - & - & - &- & - & - & - & - & -  & - & - \\
    OneIE & Cls
    & 38.5 & 52.4 & 59.3 & 61.5 & \ul{67.6} & \ul{67.4} 
    & 39.0 & 52.5 & 60.6 & 58.1 & \ul{66.5} & 66.4
    & 11.0 & 36.9 & \ul{46.7} & 48.8 & \textbf{51.8} & \textbf{53.5} \\
    Text2Event & Gen
    & 14.2 & 35.2 & 46.4 & 47.0 & 55.6 & 60.7
    & 15.7 & 38.4 & 43.9 & 46.3 & 56.5 & 62.0
    &  6.3 & 25.6 & 33.5 & 42.4 & 46.7 & 50.1 \\
    TANL & Gen 
    & 34.1 & 48.1 & 53.4 & 54.8 & 61.8 & 61.6
    & 30.3 & 50.9 & 53.1 & 55.7 & 60.8 & 61.7
    &  5.7 & 30.8 & 43.4 & 45.9 & 49.0 & 49.3 \\
    \hline
    \pipemodel{} & Gen
    & \ul{55.1} & \textbf{62.8} & \ul{63.8} & \textbf{66.1} & 64.4 & 64.4
    & \textbf{56.4} & \ul{62.5} & \ul{61.1} & \ul{62.3} & 62.5 & \textbf{67.1}
    & \textbf{32.7} & \ul{44.5} & 41.6 & \ul{50.6} & 51.1 & \textbf{53.5} \\
    \model{} & Gen
    & \textbf{55.4} & \ul{62.1} & \textbf{65.8} & \ul{65.8} & \textbf{68.3} & \textbf{68.2}
    & \ul{49.5} & \textbf{63.5} & \textbf{62.3} & \textbf{68.5} & \textbf{67.6} & \ul{66.9}
    & \ul{27.9} & \textbf{45.5} & \textbf{47.0} & \textbf{53.0} & \ul{51.7} & \textbf{53.5} \\

    \hline
    \hline
    \multicolumn{20}{|c|}{\textbf{Argument Classification F1-Score (\%)}} \\
    \hline
    \multirow{2}{*}{Model} & \multirow{2}{*}{Type} & \multicolumn{6}{c|}{ACE05-E} & \multicolumn{6}{c|}{ACE05-E$^{+}$} & \multicolumn{6}{c|}{ERE-EN} \\
    \cline{3-20}
    & & 1\% & 3\% & 5\% & 10\% & 20\% & 30\% & 1\% & 3\% & 5\% & 10\% & 20\% & 30\% & 1\% & 3\% & 5\% & 10\% & 20\% & 30\%  \\
    \hline
    BERT\_QA & Cls
    &  4.7 & 14.5 &  26.9 & 27.6 & 36.7 & 38.8 & -  & - & - & - &- & - & - & - & - & -  & - & - \\
    OneIE & Cls
    &  9.4 & 22.0 & 26.8 & 26.8 & \ul{42.7} & \ul{47.8}
    & 10.4 & 20.6 & 29.7 & 35.5 & \ul{46.7} & 48.0
    &  2.6 & 20.3 & 29.7 & 35.1 & 40.7 & \ul{43.0} \\
    Text2Event & Gen
    &  3.9 & 12.2 & 19.1 & 24.9 & 32.3 & 39.2
    &  5.7 & 16.5 & 21.3 & 26.4 & 35.2 & 42.1
    &  2.3 & 15.2 & 23.6 & 28.7 & 35.7 & 38.7 \\
    TANL & Gen 
    & 8.5 & 17.2 & 24.7 & 29.0 & 34.0 & 39.2
    & 8.6 & 22.3 & \ul{30.4} & 29.2 & 34.6 & 39.0
    & 1.4 & 20.2 & 29.5 & 30.1 & 35.6 & 36.9 \\
    \hline
    \pipemodel{} & Gen
    &  \ul{13.1} & \ul{26.1} & \ul{27.6} & \textbf{42.1} & 40.7 & 44.0
    & \ul{16.0} & \ul{26.4} & 29.9 & \ul{39.5} & 41.3 & \ul{48.5}
    & \ul{12.2} & \textbf{29.7} & \ul{31.4} & \ul{39.4} & \ul{41.9} & 42.2 \\
    \model{} & Gen
    &  \textbf{21.7} & \textbf{30.1} & \textbf{35.5} & \ul{41.6} & \textbf{46.2} & \textbf{48.7}
    & \textbf{18.7} & \textbf{34.0} & \textbf{35.7} & \textbf{43.6} &  \textbf{48.9} & \textbf{51.2}
    & \textbf{14.5} & \ul{28.9} & \textbf{33.4} & \textbf{41.7} & \textbf{42.9} & \textbf{45.5} \\
    \hline
    
\end{tabular}}
\caption{Trigger classification F1-scores and argument classification F1-scores for low-resource event extraction. Highest scores are in bold and the second best scores are underlined. ``Cls'' and ``Gen'' represent classification-based models and generation-based models, respectively. If the model is a pipelined model, then its argument predictions are based on its predicted triggers.
\model{} achieves a much better performance than other baselines. The performance gap becomes more significant for the extremely low-resource situation.}
\label{table:low_data}
\end{table*}

\begin{figure*}[t!]
    \centering
    \includegraphics[trim=0cm 0cm 0cm 0cm, clip, width=0.95\textwidth]{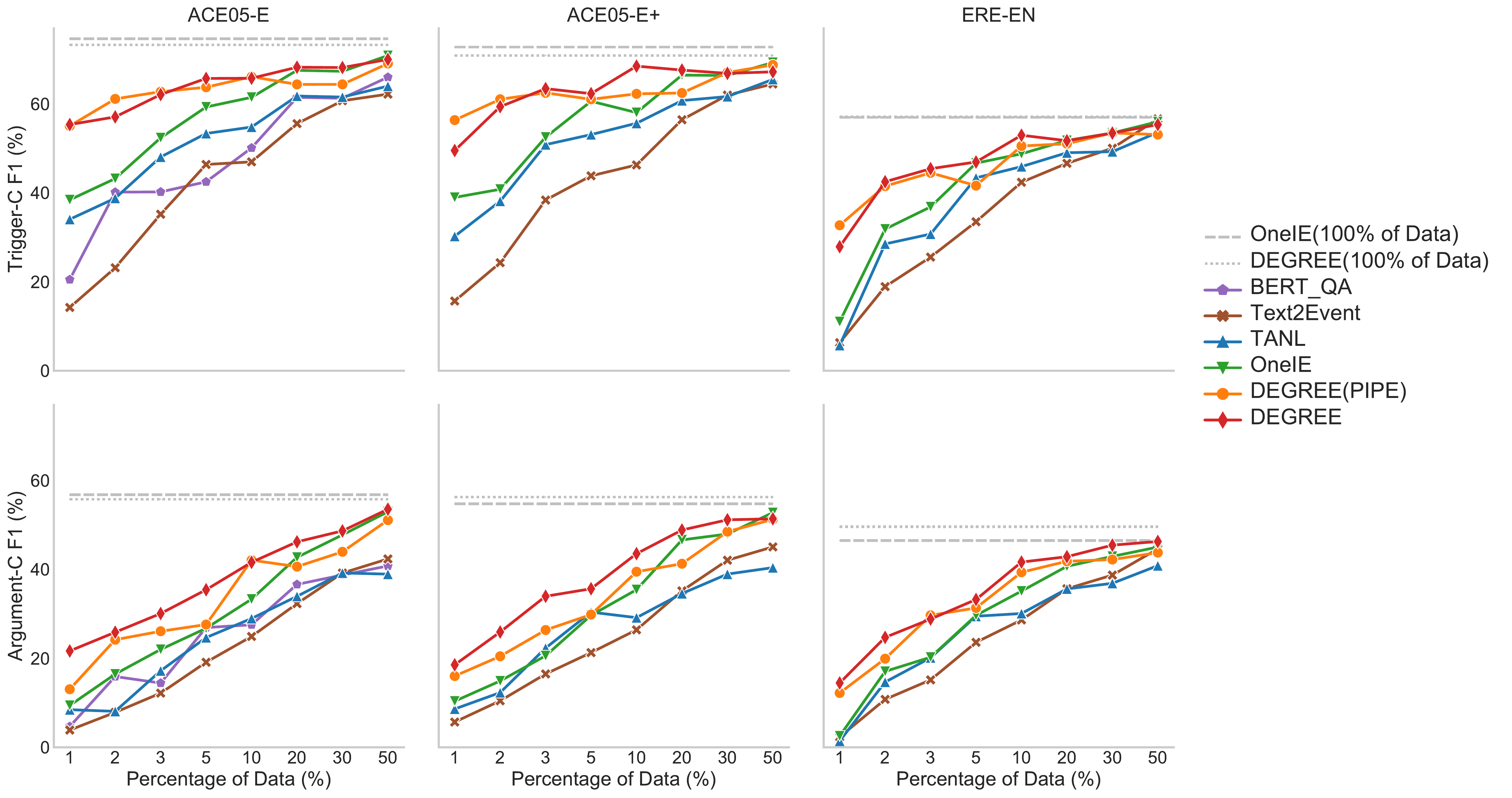}
    \caption{Trigger classification F1-scores and argument classification F1-scores for low-resource event extraction. \model{} achieves a much better performance than other baselines. The performance gap becomes more significant for the extremely low-resource situation.}
    \label{fig:low_data}
\end{figure*}

\section{Experiments}
\label{sec:low_data_result}

We conduct experiments for \emph{low-resource} event extraction to study how \model{} performs.

\subsection{Experimental Settings}

\paragraph{Datasets.}
We consider ACE 2005~\cite{doddington-etal-2004-automatic} and follow the pre-processing in \citet{Wadden19dygiepp} and \citet{Lin20oneie}, resulting in two variants: \textbf{ACE05-E} and \textbf{ACE05-E$^+$}.
Both contain 33 event types and 22 argument roles.
In addition, we consider \textbf{ERE-EN}~\cite{DBLP:conf/aclevents/SongBSRMEWKRM15} 
and adopt the pre-processing in~\citet{Lin20oneie}, which keeps 38 event types and 21 argument roles.

\paragraph{Data split for low-resource setting.}

We generate different proportions (1\%, 2\%, 3\%, 5\%, 10\%, 20\%, 30\%, and 50\%) of training data to study the influence of the size of the training set and use the original development set and test set for evaluation.
Appendix~\ref{sec:dataset} lists more details about the split generation process and the data statistics.

\paragraph{Evaluation metrics.}
We consider the same criteria in prior works~\cite{Wadden19dygiepp,Lin20oneie}.
(1) \textbf{Trigger F1-score}: an trigger is correctly identified (Tri-I) if its offset matches the gold one; it is correctly classified (Tri-C) if its event type also matches the gold one.
(2) \textbf{Argument F1-score}: an argument is correctly identified (Arg-I) if its offset and event type match the gold ones; it is correctly classified (Arg-C) if its role matches as well.

\paragraph{Compared baselines.}
We consider the following classification-based models: 
(1) \textbf{OneIE} \cite{Lin20oneie}, the current state-of-the-art (SOTA) EE model trained with designed global features.
(2) \textbf{BERT\_QA} \cite{Du20qa}, which views EE tasks as a sequence of extractive question answering problems. 
Since it learns a classifier to indicate the position of the predicted span, we view it as a classification model.
We also consider the following generation-based models:
(3) \textbf{TANL} \cite{Paolini21tacl}, which treats EE tasks as translation tasks between augmented natural languages.
(4) \textbf{Text2Event} \cite{text2event}, a sequence-to-structure model that converts the input passage to a tree-like event structure.
Note that the outputs of both generation-based baselines are not natural sentences. Therefore, it is more difficult for them to utilize the label semantics. 
All the implementation details can be found in Appendix~\ref{sec:implement}.
It is worth noting that we train OneIE with named entity annotations, as the original papers suggest, while the other models are trained without entity annotations.

\subsection{Main Results} 

Table~\ref{table:low_data} shows the trigger classification F1-scores and the argument classification F1-scores in three data sets with different proportions of training data. The results are visualized in Figure~\ref{fig:low_data}. Since our task is \textit{end-to-end} event extraction, the argument classification F1-score is the more important metric that we considered when comparing models.

From the figure and the table, we can observe that both \model{} and \pipemodel{} outperform all other baselines when using less than 10\% of the training data.
The performance gap becomes much more significant under the extremely low data situation.
For example, when only 1\% of the training data is available, \model{} and \pipemodel{} achieve more than 15 points of improvement in trigger classification F1 scores and more than 5 points in argument classification F1 scores.
This demonstrates the effectiveness of our design.
The generation-based model with carefully designed prompts is able to utilize the label semantics and the additional weakly supervised signals, thus helping learning under the low-resource regime.

Another interesting finding is that \model{} and \pipemodel{} seem to be more beneficial for predicting arguments than for predicting triggers.
For example, OneIE, the strongest baseline, requires 20\% of training data to achieve competitive performance on trigger prediction to \model{} and \pipemodel{}; however, it requires about 50\% of training data to achieve competitive performance in predicting arguments.
The reason is that the ability to capture dependencies becomes more important for argument prediction than trigger prediction since arguments are usually strongly dependent on each other compared to triggers.
Therefore, the improvements of our models for argument prediction are more significant.

Furthermore, we observe that \model{} is slightly better than \pipemodel{} under the low-resource setting. This provides empirical evidence on the benefit of jointly predicting triggers and arguments in a low-resource setting.

Finally, we perform additional experiments on few-shot and zero-shot experiments. The results can be found in Appendix~\ref{sec:fewshot-zeroshot}.

\subsection{High-Resource Event Extraction}
\label{sec:exp}

Although we focus on data-efficient learning for low-resource event extraction, to better understand the advantages and disadvantages of our model, we additionally study \model{} in the high-resource setting for controlled comparisons.

\begin{table}[t!]
\centering
\small
\setlength{\tabcolsep}{3pt}
\resizebox{1.0\columnwidth}{!}{
\begin{tabular}{|l|c|cc|cc|cc|}
    \hline
    \multirow{2}{*}{Model} & \multirow{2}{*}{Type} & \multicolumn{2}{c|}{ACE05-E} & \multicolumn{2}{c|}{ACE05-E$^{+}$} & \multicolumn{2}{c|}{ERE-EN} \\
    \cline{3-8}
    & & Tri-C &  Arg-C & Tri-C &  Arg-C & Tri-C &  Arg-C \\
    \hline
    dbRNN* & Cls
    & 69.6 & 50.1 & -  &     -  &     -  &     -  \\
    DyGIE++ & Cls
    &   70.0 &   50.0 &     -  &     -  &     -  &     -  \\
    Joint3EE* & Cls
    &   69.8 &   52.1 &     -  &     -  &     -  &     -  \\
    BERT\_QA* & Cls
    &   72.4 &   53.3 &     -  &     -  &     -  & - \\
    MQAEE* & Cls
    & 71.7 & 53.4 & -  &     -  &     -  &     -  \\
    OneIE* & Cls
    &   \textbf{74.7} &  \textbf{56.8} & \textbf{72.8} & 54.8 & 57.0  &  46.5 \\
    TANL & Gen
    &  68.4 & 47.6 &  68.6 &   46.0 &  54.7 & 43.2\\
    Text2Event* & Gen
    &  71.9 &  53.8 &  \ul{71.8}  & 54.4  &  \textbf{59.4}  & 48.3 \\
    BART-Gen* & Gen
    &   71.1 &   53.7 &     -  &     -  &     -  &     -  \\
    \hline
    \pipemodel{} & Gen
    &   72.2 &  \ul{55.8} &   71.7 &   \textbf{56.8} & \ul{57.8}  &  \textbf{50.4} \\
    \model{} & Gen
    &  \ul{73.3} &  \ul{55.8}  &  70.9 &  \ul{56.3} & 57.1 & \ul{49.6} \\
    \hline

\end{tabular}}
\caption{Results for high-resource event extraction. Highest scores are in bold and the second best scores are underlined. *We report the numbers from the original paper. \model{} has a competitive performance to the SOTA model (OneIE) and outperform other baselines.}
\label{tab:ee}
\end{table}

\begin{table}[t!]
\centering
\small
\setlength{\tabcolsep}{3pt}
\resizebox{1.0\columnwidth}{!}{
\begin{tabular}{|l|c|cc|cc|cc|}
    \hline
    \multirow{2}{*}{Model} & \multirow{2}{*}{Type} & \multicolumn{2}{c|}{ACE05-E} & \multicolumn{2}{c|}{ACE05-E$^{+}$} & \multicolumn{2}{c|}{ERE-EN} \\
    \cline{3-8}
    & &  Arg-I &  Arg-C &  Arg-I &  Arg-C &  Arg-I &  Arg-C \\
    \hline
    DyGIE++ & Cls
    &   66.2 &   60.7 &     -  &     -  &     -  &     -  \\
    BERT\_QA* & Cls
    &   68.2 &   65.4 &     -  &     -  &     -  &     -  \\
    OneIE & Cls
    &   73.2 &   69.3 &  73.3 &   70.6 &  75.3  & 70.0  \\
    TANL & Gen
    &   65.9 &   61.0 &     66.3  & 62.3  &   75.6  &   69.6  \\
    BART-Gen* & Gen
    &   69.9 &   66.7 &     -  &     -  &     -  &     -  \\
    \hline
    \eaemodel{} & Gen
    &   \textbf{76.0} &   \textbf{73.5} &   \textbf{75.2} &   \textbf{73.0} &  \textbf{80.2}  &   \textbf{76.3} \\
    \hline

\end{tabular}}
\caption{Results for high-resource event argument extraction. Models predict arguments based on the given gold triggers. Best scores are in bold. *We report the numbers from the original paper. \eaemodel{} achieves a new state-of-the-art performance on event argument extraction.}
\label{tab:eae}
\end{table}

\paragraph{Compared baselines.}
In addition to the EE models mentioned above:
\textbf{OneIE} \cite{Lin20oneie}, 
\textbf{BERT\_QA} \cite{Du20qa},  \textbf{TANL}~\cite{Paolini21tacl}, 
and \textbf{Text2Event} \cite{text2event}, 
we also consider the following baselines focusing on the high-resource setting.
\textbf{dbRNN} \cite{Sha18dbrnn} is classification-based model that adds dependency bridges for event extraction.
\textbf{DyGIE++} \cite{Wadden19dygiepp} is a classification-based model with span graph propagation technique.
\textbf{Joint3EE} \cite{Nguyen19joint3ee} is a classification-based model jointly trained with annotations of entity, trigger, and argument.
\textbf{MQAEE}~\cite{li-etal-2020-event} converts EE to a series of question answering problems for argument extraction . \textbf{BART-Gen}~\cite{li2021document} is a generation-based model focusing on \emph{only} event argument extraction.\footnote{We follow the original paper and use \textsc{TapKey} as their event detection model.}
Appendix~\ref{sec:implement} shows the implementation details for the baselines.

\begin{table}[t!]
\centering
\small
\resizebox{\columnwidth}{!}{
\setlength{\tabcolsep}{4pt}
\begin{tabular}{|l|cc|cc|}
    \hline
    \multirow{2}{*}{Model}       & \multicolumn{2}{c|}{10\% Data} & \multicolumn{2}{c|}{100\% Data} \\
    \cline{2-5}
    & Tri-I         & Tri-C         & Tri-I          & Tri-C         \\
    \hline
    Full \edmodel{}   & \textbf{69.3} & \textbf{66.1}    &\textbf{75.4}&   \textbf{72.2} \\
    - w/o \Eventdef{} & 67.9 & 64.4    &   73.5 &   70.1 \\
    - w/o \edtemp{}   & 68.8 & 65.8    &   74.0 &   70.5 \\
    - w/o \Eventkey{} & 68.2 & 64.0    &   73.5 &   69.1 \\
    - only \Eventdef{}  & 66.3 & 63.5 &   72.6 &   68.9 \\
    - only \Eventkey{}  & 69.2 & 63.8 &   70.8 &   66.2 \\
    \hline

\end{tabular}}
\caption{Ablation study for the components in the prompt on event detection with ACE05-E.}
\label{tab:input_ed}
\end{table}

\paragraph{Results for event extraction.}
\label{sec:exp_ee}

Table~\ref{tab:ee} shows the results of high-resource event extraction.
In terms of trigger predictions (Tri-C), \model{} and \pipemodel{} outperforms all the baselines except for OneIE, the current state-of-the-art model.
For argument predictions (Arg-C), our models have slightly better performance than OneIE in two out of the three datasets.
When enough training examples are available, models can learn more sophisticated features from data, which do not necessarily follow the learned dependencies.
Therefore, the advantage of \model{} over \pipemodel{} becomes less obvious.
This result justifies our hypothesis that \model{} has better performance for the \textit{low-resource setting} because of its ability to better capture dependencies.

\paragraph{Results for event argument extraction.} 
In Table~\ref{tab:eae}, we additionally study the performance for event argument extraction task, where the model makes argument predictions \textit{with the gold trigger provided}.
Interestingly, \eaemodel{} achieves pretty strong performance and outperforms other baselines with a large margin.
Combining the results in Table~\ref{tab:ee}, we hypothesize that event argument extraction is a more challenging task than event trigger detection and it requires more training examples to learn well.
Hence, our proposed model, which takes the advantage of using label semantics to better capture dependencies, achieves a new state-of-the-art for event argument extraction.

\section{Ablation Studies}
\label{sec:ablation}
In this section, we present comprehensive ablation studies to justify our design.
To better understand the contribution of each component in the designed prompt and their effects on the different tasks, we ablate \eaemodel{} and \edmodel{} for both low-resource and high-resource situations. 

\paragraph{Impacts of components in prompts.}
Table~\ref{tab:input_ed} lists the performance changes when removing the components in the prompts for event detection on ACE05-E. The performance decreases whenever removing any one of \eventdef{}, \eventkey{}, and \edtemp{}.
The results suggest that three components are all necessary.

Table~\ref{tab:input_eae} demonstrates how different components in prompts affect the performance of event argument extraction on ACE05-E. Removing any one of \eventdef{}, \querytrig{}, and \eaetemp{} leads to performance drops, which validates their necessity.
We observe that \querytrig{} plays the most important role among the three and when less training data is given, the superiority of leveraging any of these weakly-supervised signal becomes more obvious.



\begin{table}[t!]
\centering
\small
\resizebox{\columnwidth}{!}{
\setlength{\tabcolsep}{3.5pt}
\begin{tabular}{|l|cc|cc|}
\hline
\multirow{2}{*}{Model}       & \multicolumn{2}{c|}{10\% Data} & \multicolumn{2}{c|}{100\% Data} \\
\cline{2-5}
& Arg-I         & Arg-C         & Arg-I          & Arg-C         \\
\hline
Full \eaemodel{}   &  \textbf{63.3} &  \textbf{57.3}  & \textbf{76.0}  & \textbf{73.5} \\
- w/o \Eventdef{}  &  60.3     &  54.4     & 74.5           & 71.1          \\
- w/o \eaetemp{}   &   57.0   &   51.9     & 73.8           & 70.4 \\
- w/o \Querytrig{} &   55.2   &  49.9    & 71.4           & 69.0          \\
- only \Querytrig{} &   51.9  &   48.1 &   71.2  &   69.4  \\
- only \eaetemp{}   &   51.2  &   46.9 &   71.4  &   68.6  \\
- only \Eventdef{}  &   46.7  &   42.3 &   71.4  &   68.2 \\
    \hline

\end{tabular}}
\caption{Ablation study for the components in the prompt on event argument extraction with ACE05-E.}
\label{tab:input_eae}
\end{table}

\paragraph{Effects of different template designs.}
To verify the importance of using natural sentences as outputs, we study three variants of \eaetemps{}:
\begin{itemize}[topsep=3pt, itemsep=-1pt, leftmargin=12pt]
    \item \textbf{Natural sentence.} Our proposed templates described in Section~\ref{sec:method}, e.g., ``\textit{somebody was born in somewhere.}'',
    where \textit{``somebody''} and \textit{``somewhere''} are placeholders that can be replaced by the corresponding arguments.
    \item \textbf{Natural sentence with special tokens.}
    It is similar to the natural sentence one except for using role-specific special tokens instead of ``some-'' words.
    For example, ``\textit{$<$Person$>$ was born in $<$Place$>$.}''
    We consider this to study the \emph{label semantics} of roles.
    \item \textbf{HTML-like sentence with special tokens.} To study the importance of using natural sentence, we also consider HTML-like sentence, e.g., ``\textit{$<$Person$>$ $<$/Person$>$ $<$Place$>$ $<$/Place$>$}''. The model aims to put argument predictions between the corresponding HTML tags.
\end{itemize}

The results of all variants of \eaetemps{} on ACE05-E are shown in Table~\ref{tab:output_eae}. 
We notice that writing templates in a natural language style get better performance, especially when only a few data is available (10\% of data). This shows our design's capability to leverage pre-trained knowledge in the generation process.
Additionally, there are over 1 F1 score performance drops when replacing natural language placeholders with special tokens. This confirms that leveraging label semantics for different roles is beneficial.

\begin{table}[t!]
\centering
\small
\resizebox{\columnwidth}{!}{
\setlength{\tabcolsep}{3pt}
\begin{tabular}{|l|cc|cc|}
    \hline
\multirow{2}{*}{Model}       & \multicolumn{2}{c|}{10\% Data} & \multicolumn{2}{c|}{100\% Data} \\
\cline{2-5}
    & Arg-I         & Arg-C         & Arg-I          & Arg-C         \\
    \hline
    OneIE
    & 48.3 & 45.4 & 73.2 & 69.3 \\
    BART-Gen
    & - & - & 69.9 & 66.7 \\
    \hline
    Natural sentence     &\textbf{63.3}   &  \textbf{57.3}            &   \textbf{76.0} &   \textbf{73.5} \\
    Natural sentence w/ special tokens  & 59.8 & 55.5 &   74.7 &   72.3 \\
    HTML-like sentence w/ special tokens & 60.8 & 51.9 &   74.6 &   71.4 \\
    \hline

\end{tabular}}
\caption{Performances of \eaemodel{} on ACE05-E with different types of templates.}
\label{tab:output_eae}
\end{table}



\begin{table}[t!]
\centering
\small
\resizebox{\columnwidth}{!}{
\setlength{\tabcolsep}{3pt}
\begin{tabular}{|l|cc|cc|}
    \hline
\multirow{2}{*}{Model}       & \multicolumn{2}{c|}{10\% Data} & \multicolumn{2}{c|}{100\% Data} \\
\cline{2-5}
    & Arg-I         & Arg-C         & Arg-I          & Arg-C         \\
    \hline
    OneIE
    & 48.3 & 45.4 & 73.2 & 69.3 \\
    BART-Gen
    & - & - & 69.9 & 66.7 \\
    \hline
    \eaemodel{}   &  63.3 &  \textbf{57.3}  & \textbf{76.0}  & \textbf{73.5} \\
    \eaemodel{} + variant template 1 
    & 61.6 & 55.5 & 73.4 & 70.4 \\
    \eaemodel{} + variant template 2 
    & \textbf{63.9} & 56.9 & 75.5 & 72.5 \\
    \hline

\end{tabular}}
\caption{Study on the effect of different template constructing rules. Experiments is conducted on ACE05-E.}
\vspace{-0.5em}
\label{tab:sens}
\end{table}

\paragraph{Sensitivity to template design.}
Finally, we study how sensitive our model is to the template.
In addition to the original design of templates for event argument extraction, we compose other two sets of templates with different constructing rules (e.g., different word choices and different orders of roles).
Table~\ref{tab:sens} shows the results of using different sets of templates.
We observe a performance fluctuation when using different templates, which indicates that the quality of templates does affect the performance to a certain degree.
Therefore, we need to be cautious when designing templates.
However, even though our model could be sensitive to the template design, it still outperforms OneIE and BART-Gen, which are the best classification-based model and the best generation-based baseline, respectively.
\section{Related Work}
\label{sec:related_work}
\paragraph{Fully supervised event extraction.}
Event extraction has been studied for over a decade~\cite{ahn-2006-stages, ji-grishman-2008-refining} and most traditional event extraction works follow the fully supervised setting~\cite{Nguyen16jrnn, Sha18dbrnn, Nguyen19joint3ee, yang-etal-2019-exploring, Lin20oneie, Liu2020rceeer, li-etal-2020-event}.
Many of them use classification-based models and use pipeline-style frameworks to extract events~\cite{Nguyen16jrnn, yang-etal-2019-exploring, Wadden19dygiepp}. To better leverage shared knowledge in event triggers and arguments, some works propose incorporating global features to jointly decide triggers and arguments~\cite{Lin20oneie, Li13jointbeam, yang-mitchell-2016-joint}.

Recently, few generation-based event extraction models have been proposed~\cite{Paolini21tacl, DBLP:conf/emnlp/HuangTP21, acl2022xgear, li2021document}. TANL \cite{Paolini21tacl} treats event extraction as translation tasks between augmented natural languages. Their predicted target---augmented language embed labels into the input passage via using brackets and vertical bar symbols. TempGen~\cite{DBLP:conf/emnlp/HuangTP21} is a template-based role-filler entity extraction model, which generate outputs that fill role entities into non-natural templated sequences. The output sequence designs of TANL and TempGen hinder the models from fully leveraging label semantics, unlike \model{} that generates natural sentences.
BART-Gen~\cite{li2021document} is also a generation-based model focusing on document-level event argument extraction. They solve event extraction with a pipeline, which prevents knowledge sharing across subtasks. All these fully supervised methods can achieve substantial performance with a large amount of annotated data. However, their designs are not specific for low-resource scenarios, hence, these models can not enjoy all the benefits that \model{} obtains for low-resource event extraction at the same time, as we mentioned in Section~\ref{sec:intro}.

\paragraph{Low-resource event extraction.}
It has been a growing interest in event extraction in a scenario with less data. \citet{Liu2020rceeer} uses a machine reading comprehension formulation to conduct event extraction in a low-resource regime. Text2Event~\cite{text2event}, a sequence-to-structure generation paradigm, first presents events in a linearized format, and then trains a generative model to generate the linearized event sequence. Text2Event's unnatural output format hinders the model from fully leveraging pre-trained knowledge. Hence, their model falls short on the cases with only extremely low data being available (as shown in Section~\ref{sec:low_data_result}).

Another thread of works are using meta-learning to deal with the less label challenge~\cite{DBLP:conf/wsdm/DengZKZZC20, DBLP:conf/acl/ShenWQLHB21, DBLP:conf/acl/CongCYLWW21}. However, their methods can only be applied to event detection, which differs from our main focus on studying end-to-end event extraction.
\section{Conclusion \& Future Work}
\label{sec:conclusion}

In this paper, we present \model{}, a data-efficient generation-based event extraction model.
\model{} requires less training data because it better utilizes label semantics as well as weakly-supervised information, and captures better dependencies by jointly predicting triggers and arguments.
Our experimental results and ablation studies show the superiority of \model{} for low-resource event extraction.

\model{} assumes that some weakly-supervised information (the description of events, similar keywords, and human-written templates) is accessible or not expensive for the users to craft.
This assumption may holds for most situations. We leave the automation of template construction for future work, which can further ease the needed efforts when deploying \model{} in a large-scale corpus. 

\section*{Acknowledgments}
We thank anonymous reviewers for their helpful feedback. 
We thank the UCLA PLUSLab and UCLA-NLP group members for their initial review and feedback for an earlier version of the paper. 
This work is supported in part by the Intelligence Advanced
Research Projects Activity (IARPA), via Contract
No.\ 2019-19051600007, and research awards sponsored by CISCO and Google.

\section*{Ethics Considerations}

\model{} fine-tunes the pre-trained generative language model \cite{lewis-etal-2020-bart}.
Therefore, the generated output is potentially affected by the corpus for pre-training.
Although with a low possibility, it is possible for our model to accidentally generate some malicious, counterfactual, and biased sentences, which may cause ethics concerns.
We suggest carefully examining those potential issues before deploying the model in any real-world applications.

\bibliography{acl}
\bibliographystyle{acl_natbib}

\clearpage
\clearpage
\appendix

\section{EAE Template Constructing}
\label{sec:eaetemp_list}

Our strategy to create an \eaetemp{} is first identifying all valid argument roles for the event type,\footnote{The valid roles for each event type are predefined in the event ontology for each dataset, or can be decided by the user of interest.} such as \textit{Attacker}, \textit{Target}, \textit{Instrument}, and \textit{Place} roles.
Then, for each argument role, according to the semantics of the role type, we select natural and fluent words to form its placeholder (e.g., \textit{\ul{some way}} for \textit{Instrument}).
This design aims to provide a simple way to help the model learn both the roles' label semantics and the \textit{event structure}.
Finally, we create a natural language sentence that connects all these placeholders. 
Notice that we try to keep the template as simple and short as possible.
Table \ref{tab:temp-ace} lists all designed \eaetemps{} for ACE05-E and ACE05-E$^+$. The \eaetemps{} of ERE-EN can be found in Table~\ref{tab:temp-ere}.

\begin{table*}[t]
\centering
\small
\setlength{\tabcolsep}{5pt}
\begin{tabular}{|x{.3\textwidth}|x{.65\textwidth}|}
    \hline
    \textbf{Event Type} & \textbf{EAE Template} \\
    \hline
Life:Be-Born & somebody was born in somewhere. \\
\hline
Life:Marry & somebody got married in somewhere. \\
\hline
Life:Divorce & somebody divorced in somewhere. \\
\hline
Life:Injure & somebody or some organization led to some victim injured by some way in somewhere. \\
\hline
Life:Die & somebody or some organization led to some victim died by some way in somewhere. \\
\hline
Movement:Transport & something was sent to somewhere from some place by some vehicle. somebody or some organization was responsible for the transport. \\
\hline
Transaction:Transfer-Ownership & someone got something from some seller in somewhere. \\
\hline
Transaction:Transfer-Money & someone paid some other in somewhere. \\
\hline
Business:Start-Org & somebody or some organization launched some organzation in somewhere. \\
\hline
Business:Merge-Org & some organzation was merged. \\
\hline
Business:Declare-Bankruptcy & some organzation declared bankruptcy. \\
\hline
Business:End-Org & some organzation dissolved. \\
\hline
Conflict:Attack & some attacker attacked some facility, someone, or some organization by some way in somewhere. \\
\hline
Conflict:Demonstrate & some people or some organization protest at somewhere. \\
\hline
Contact:Meet & some people or some organization met at somewhere. \\
\hline
Contact:Phone-Write & some people or some organization called or texted messages at somewhere. \\
\hline
Personnel:Start-Position & somebody got new job and was hired by some people or some organization in somewhere. \\
\hline
Personnel:End-Position & somebody stopped working for some people or some organization at somewhere. \\
\hline
Personnel:Nominate & somebody was nominated by somebody or some organization to do a job. \\
\hline
Personnel:Elect & somebody was elected a position, and the election was voted by some people or some organization in somewhere. \\
\hline
Justice:Arrest-Jail & somebody was sent to jailed or arrested by somebody or some organization in somewhere. \\
\hline
Justice:Release-Parole & somebody was released by some people or some organization from somewhere. \\
\hline
Justice:Trial-Hearing & somebody, prosecuted by some other, faced a trial in somewhere. The hearing was judged by some adjudicator. \\
\hline
Justice:Charge-Indict & somebody was charged by some other in somewhere. The adjudication was judged by some adjudicator. \\
\hline
Justice:Sue & somebody was sued by some other in somewhere. The adjudication was judged by some adjudicator. \\
\hline
Justice:Convict & somebody was convicted of a crime in somewhere. The adjudication was judged by some adjudicator. \\
\hline
Justice:Sentence & somebody was sentenced to punishment in somewhere. The adjudication was judged by some adjudicator. \\
\hline
Justice:Fine & some people or some organization in somewhere was ordered by some adjudicator to pay a fine. \\
\hline
Justice:Execute & somebody was executed by somebody or some organization at somewhere. \\
\hline
Justice:Extradite & somebody was extradicted to somewhere from some place. somebody or some organization was responsible for the extradition. \\
\hline
Justice:Acquit & somebody was acquitted of the charges by some adjudicator. \\
\hline
Justice:Pardon & somebody received a pardon from some adjudicator. \\
\hline
Justice:Appeal & some other in somewhere appealed the adjudication from some adjudicator. \\
\hline
\end{tabular}
\caption{All \eaetemps{} for ACE05-E and ACE05-E$^+$.}
\label{tab:temp-ace}
\end{table*}

\begin{table*}[t]
\centering
\small
\setlength{\tabcolsep}{5pt}
\begin{tabular}{|x{.3\textwidth}|x{.65\textwidth}|}
    \hline
    \textbf{Event Type} & \textbf{EAE Template} \\
    \hline
Life:Be-Born & somebody was born in somewhere. \\
\hline
Life:Marry & somebody got married in somewhere. \\
\hline
Life:Divorce & somebody divorced in somewhere. \\
\hline
Life:Injure & somebody or some organization led to some victim injured by some way in somewhere. \\
\hline
Life:Die & somebody or some organization led to some victim died by some way in somewhere. \\
\hline
Movement:Transport-Person & somebody was moved to somewhere from some place by some way. somebody or some organization was responsible for the movement. \\
\hline
Movement:Transport-Artifact & something was sent to somewhere from some place. somebody or some organization was responsible for the transport. \\
\hline
Business:Start-Org & somebody or some organization launched some organzation in somewhere. \\
\hline
Business:Merge-Org & some organzation was merged. \\
\hline
Business:Declare-Bankruptcy & some organzation declared bankruptcy. \\
\hline
Business:End-Org & some organzation dissolved. \\
\hline
Conflict:Attack & some attacker attacked some facility, someone, or some organization by some way in somewhere. \\
\hline
Conflict:Demonstrate & some people or some organization protest at somewhere. \\
\hline
Contact:Meet & some people or some organization met at somewhere. \\
\hline
Contact:Correspondence & some people or some organization contacted each other at somewhere. \\
\hline
Contact:Broadcast & some people or some organization made announcement to some publicity at somewhere. \\
\hline
Contact:Contact & some people or some organization talked to each other at somewhere. \\
\hline
Manufacture:Artifact & something was built by somebody or some organization in somewhere. \\
\hline
Personnel:Start-Position & somebody got new job and was hired by some people or some organization in somewhere. \\
\hline
Personnel:End-Position & somebody stopped working for some people or some organization at somewhere. \\
\hline
Personnel:Nominate & somebody was nominated by somebody or some organization to do a job. \\
\hline
Personnel:Elect & somebody was elected a position, and the election was voted by somebody or some organization in somewhere. \\
\hline
Transaction:Transfer-Ownership & The ownership of something from someone was transferred to some other at somewhere. \\
\hline
Transaction:Transfer-Money & someone paid some other in somewhere. \\
\hline
Transaction:Transaction & someone give some things to some other in somewhere. \\
\hline
Justice:Arrest-Jail & somebody was sent to jailed or arrested by somebody or some organization in somewhere. \\
\hline
Justice:Release-Parole & somebody was released by somebody or some organization from somewhere. \\
\hline
Justice:Trial-Hearing & somebody, prosecuted by some other, faced a trial in somewhere. The hearing was judged by some adjudicator. \\
\hline
Justice:Charge-Indict & somebody was charged by some other in somewhere. The adjudication was judged by some adjudicator. \\
\hline
Justice:Sue & somebody was sued by some other in somewhere. The adjudication was judged by some adjudicator. \\
\hline
Justice:Convict & somebody was convicted of a crime in somewhere. The adjudication was judged by some adjudicator. \\
\hline
Justice:Sentence & somebody was sentenced to punishment in somewhere. The adjudication was judged by some adjudicator. \\
\hline
Justice:Fine & some people or some organization in somewhere was ordered by some adjudicator to pay a fine. \\
\hline
Justice:Execute & somebody was executed by somebody or some organization at somewhere. \\
\hline
Justice:Extradite & somebody was extradicted to somewhere from some place. somebody or some organization was responsible for the extradition. \\
\hline
Justice:Acquit & somebody was acquitted of the charges by some adjudicator. \\
\hline
Justice:Pardon & somebody received a pardon from some adjudicator. \\
\hline
Justice:Appeal & somebody in somewhere appealed the adjudication from some adjudicator. \\
\hline

\end{tabular}
\caption{All \eaetemps{} for ERE-EN.}
\label{tab:temp-ere}
\end{table*}

\section{Training Details of Proposed Model}
\label{sec:training}
Given a passage, its annotated event types are consider as positive event types.
During training, we additionally sample $m$ event types that are not related to the passage as the negative examples, where $m$ is a hyper-parameter.
In our experiments, $m$ is usually set to 13 or 15.

For all of \model{}, \edmodel{}, and \eaemodel{}, we fine-tune the pre-trained BART-large \cite{lewis-etal-2020-bart} with Huggingface package \cite{huggingface}.
The number of parameters is around 406 millions.
We train \model{} with our machine that equips 128 AMD EPYC 7452 32-Core Processor, 4 NVIDIA A100 GPUs, and 792G RAM.
We consider AdamW optimizer~\cite{DBLP:conf/iclr/LoshchilovH19} with learning rate set to $10^{-5}$ and the weight decay set to $10^{-5}$. 
We set the batch size to 6 for \eaemodel{} and 32 for \edmodel{} and \model{}. 
The number of training epochs is 45. 
It takes around 2 hours, 18 hours, 22 hours to train \eaemodel{}, \edmodel{}, and \model{}, respectively.

We do hyper-parameter search on $m$, the number of negative examples, from $\{ 3, 5, 7, 10, 13, 15, 18, 21\}$, and our preliminary trials shows that $m$ less than 10 are usually less useful. For the learning rate and the weight decay, we tune it based on our preliminary experiment for event argument extraction from $\{ 10^{-5}, 10^{-4}\}$, while they are both fixed to $10^{-5}$ for all the experiments.

\section{Datasets}
\label{sec:dataset}

We consider ACE 2005\footnote{\url{https://catalog.ldc.upenn.edu/LDC2006T06}} \cite{doddington-etal-2004-automatic} and ERE\footnote{\url{https://catalog.ldc.upenn.edu/LDC2020T19}} \cite{DBLP:conf/aclevents/SongBSRMEWKRM15}.
Both consider \emph{LDC User Agreement for Non-Members}\footnote{\url{https://catalog.ldc.upenn.edu/license/ldc-non-members-agreement.pdf}} as the licenses.
Both datasets are created for entity, relation, and event extraction while our focus is only event extraction in this paper.
In the original ACE 2005 dataset, it contains data for English, Chinese, and Arabic and we only take the English data for our experiment.
In the original ERE dataset, it contains data for English, and Chinese and we only take the English data for our experiment as well.

Because both datasets contain event like \emph{Justice:Execute} and \emph{Life:Die}, it is possible that some offensive words (e.g., killed) would appear in the passage.
Also, some real names may appear in the passage as well (e.g., Palestinian president, Mahmoud Abbas).
How to accurately identify these kinds of information is part of the goal of the task.
Therefore, we do not take any changes on the datasets for protecting or anonymizing.

We split the training data based on documents, which is a more realistic setup compared to splitting data by instance.
Table~\ref{tab:stats} lists the statistics of ACE05-E, ACE05-E$^+$, and ERE-EN.
Specifically, we try to make each proportion of data contain as many event types as possible.

\begin{table*}[ht]
\centering
\small
\setlength{\tabcolsep}{5pt}
\begin{tabular}{|l|l|cccccc|}
    \hline
    Dataset & Split & \#Docs& \#Sents & \#Events & \#Event Types & \#Args & \#Arg Types\\
    \hline
    \multirow{11}{*}{ACE05-E}
    & Train (full) & 529 & 17172 & 4202 & 33 & 4859 & 22 \\
    & Train (1\%) & 5 & 103 & 47 & 14 & 65 & 16 \\
    & Train (2\%) & 10 & 250 & 77 & 17 & 104 & 16 \\
    & Train (3\%) & 15 & 451& 119& 23& 153& 17 \\
    & Train (5\%) & 25& 649& 212& 27& 228& 21 \\
    & Train (10\%) & 50& 1688& 412& 28& 461& 21 \\
    & Train (20\%) &110& 3467& 823& 33& 936& 22 \\
    & Train (30\%) &160& 5429& 1368& 33& 1621& 22 \\
    & Train (50\%) &260& 8985& 2114& 33& 2426& 22 \\
    \cline{2-8}
    & Dev   &   28& 923& 450& 21& 605& 22 \\
    & Test  &   40& 832& 403& 31& 576& 20 \\
    \hline\hline
    \multirow{11}{*}{ACE05-E$^+$}
    & Train (full) & 529& 19216& 4419& 33& 6607& 22 \\
    & Train (1\%) & 5& 92& 49& 15& 75& 16 \\
    & Train (2\%) & 10& 243& 82& 19& 129& 16 \\
    & Train (3\%) & 15& 434& 124& 24& 203& 19 \\
    & Train (5\%) & 25& 628& 219& 27& 297& 21 \\
    & Train (10\%) & 50& 1915& 428& 29& 629& 21 \\
    & Train (20\%) &110& 3834& 878& 33& 1284& 22 \\
    & Train (30\%) &160& 6159& 1445& 33& 2212& 22 \\
    & Train (50\%) &260& 10104& 2231& 33& 3293& 22 \\
    \cline{2-8}
    & Dev   &   28& 901& 468& 22& 759& 22 \\
    & Test  &   40& 676& 424& 31& 689& 21 \\
    \hline
    \hline
    \multirow{11}{*}{ERE-EN}
    & Train (full) &396& 14736& 6208& 38& 8924& 21 \\
    & Train (1\%) & 4& 109& 61& 14& 78& 16 \\
    & Train (2\%) & 8& 228& 128& 21& 183& 19 \\
    & Train (3\%) & 12& 419& 179& 26& 272& 19 \\
    & Train (5\%) & 20& 701& 437& 31& 640& 21 \\
    & Train (10\%) & 40& 1536& 618& 37& 908& 21 \\
    & Train (20\%) & 80& 2848& 1231& 38& 1656& 21 \\
    & Train (30\%) &120& 4382& 1843& 38& 2632& 21 \\
    & Train (50\%) &200& 7690& 3138& 38& 4441& 21 \\
    \cline{2-8}
    & Dev   &   31& 1209& 525& 34& 730& 21 \\
    & Test  &   31& 1163& 551& 33& 822& 21 \\
    \hline

\end{tabular}
\caption{Dataset statistics. Our experiments are conducted in sentences, which were split from documents. In the table, ``\#Docs'' means the number of documents; ``\#Sents'' means the number of sentences, ``\#Events'' means the number of events; ``\#Event Types'' means the number of event types in total; ``\#Args'' means the number of argument in total; ``\#Arg Types'' means the number of argument role types in total.}
\label{tab:stats}
\end{table*}

\section{Implementation Details}
\label{sec:implement}
This section describes the implementation details for all baselines we use. We run the experiments with three different random seeds and report the best value.

\begin{itemize}[topsep=3pt, itemsep=-2pt, leftmargin=13pt]
    \item \textbf{DyGIE++}: we use their released pre-trained model\footnote{\url{https://github.com/dwadden/dygiepp}} for evaluation.
    \item \textbf{OneIE}: we use their provided code\footnote{\url{http://blender.cs.illinois.edu/software/oneie/}} to train the model with default parameters.
    \item \textbf{BERT\_QA}: we use their provided code\footnote{\url{https://github.com/xinyadu/eeqa}} to train the model with default parameters.
    \item \textbf{TANL}: we use their provided code\footnote{\url{https://github.com/amazon-research/tanl}} to train the model. We conduct the experiments with two variations: (1) using their default parameters, and (2) using their default parameters but with more training epochs. We observe that the second variant works better. As a result, we report the number obtained from the second setting.
    \item \textbf{Text2Event}: we use their official code\footnote{\url{https://github.com/luyaojie/Text2Event}} to train the model with the provided parameter setting.
    \item \textbf{dbRNN:} we directly report the experimental results from their paper.
    \item \textbf{Joint3EE:} we directly report the experimental results from their paper.
    \item \textbf{MQAEE:} we directly report the experimental results from their paper.
    \item \textbf{BART-Gen}: we report the experimental results from their released appendix.\footnote{\url{https://github.com/raspberryice/gen-arg/blob/main/NAACL_2021_Appendix.pdf}}
\end{itemize}

\clearpage

\section{Few-Shot and Zero-Shot Event Extraction}
\label{sec:fewshot-zeroshot}

In order to further test our models' generaliability, we additionally conduct zero-shot and few-shot experiments on the ACE05-E dataset with \edmodel{} and \eaemodel{}. 

\paragraph{Settings.}
We first select the top $n$ common event types as ``seen'' types and use the rest as ``unseen/rare'' types, where the top common types are listed in Table~\ref{tab:common_event}. 
To simulate a zero-shot scenario, we remove all events with ``unseen/rare'' types from the training data. To simulate a few-shot scenario, we keep only $k$ event examples for each  ``unseen/rare'' type (denoted as $k$-shot). 
During the evaluation, we calculate micro F1-scores only for these ``unseen/rare'' types. 

\begin{table}[!ht]
\centering
\small
\setlength{\tabcolsep}{3pt}
\begin{tabular}{|y{.05\columnwidth}|x{.85\columnwidth}|}
    \hline
    n &  Seen Event Types for Training/Development \\
    \hline
    5  &  Conflict:Attack, Movement:Transport, Life:Die, Contact:Meet, Personnel:Elect   \\
    \hline
    10 &  \makecell[l]{Conflict:Attack, Movement:Transport, Life:Die, \\ 
                    Contact:Meet, Personnel:Elect, Life:Injure, \\
                    Personnel:End-Position, Justice:Trial-Hearing, \\ 
                    Contact:Phone-Write, Transaction:Transfer-Money} \\
    \hline
\end{tabular}
\caption{Common event types in ACE05-E.}
\label{tab:common_event}
\end{table}

\paragraph{Compared baselines.}
We consider the following baselines:
(1) \textbf{BERT\_QA}~\cite{Du20qa}
(2) \textbf{OneIE}~\cite{Lin20oneie}
(3) \textbf{Matching baseline}, a proposed baseline that makes trigger predictions by performing string matching between the input passage and the \eventkey{}. (4) \textbf{Lemmatization baseline}, another proposed baseline that performs string matching on lemmatized input passage and the \eventkey{}.
(Note: (3) and (4) are baselines only for event detection tasks.) 

\begin{figure*}[t!]
    \centering
    \begin{subfigure}[b]{0.99\textwidth}
    \centering
    \includegraphics[trim=0cm 0cm 0cm 0cm, clip, width=1.0\textwidth]{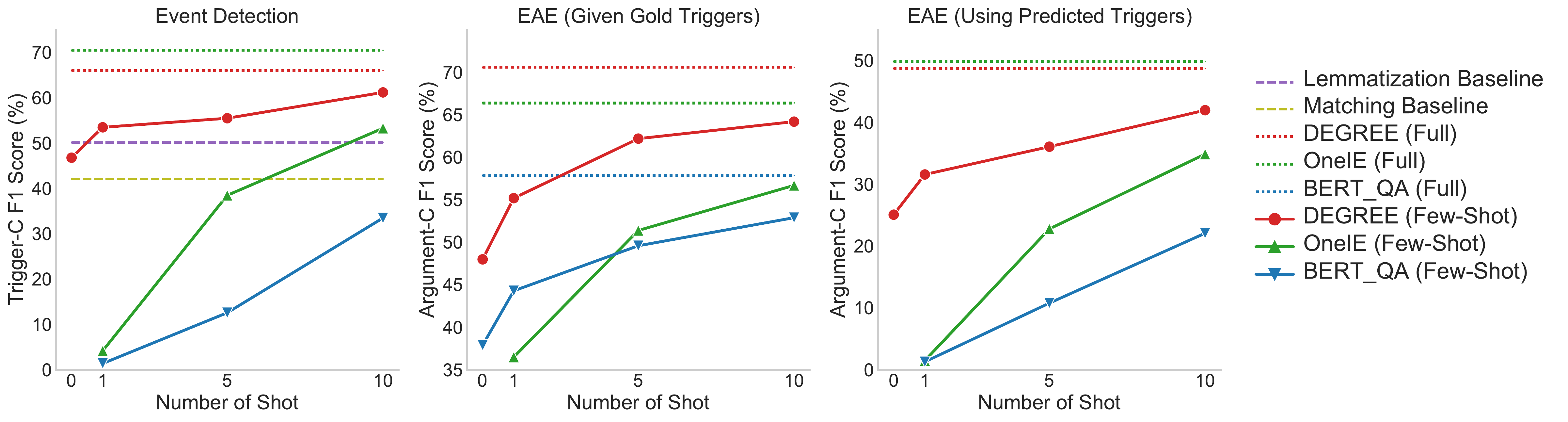}
    \caption{Results for top common 5 event types.}
    \end{subfigure}
    \begin{subfigure}[b]{0.99\textwidth}
    \par\bigskip 
    \includegraphics[trim=0cm 0cm 0cm 0cm, clip, width=1.0\textwidth]{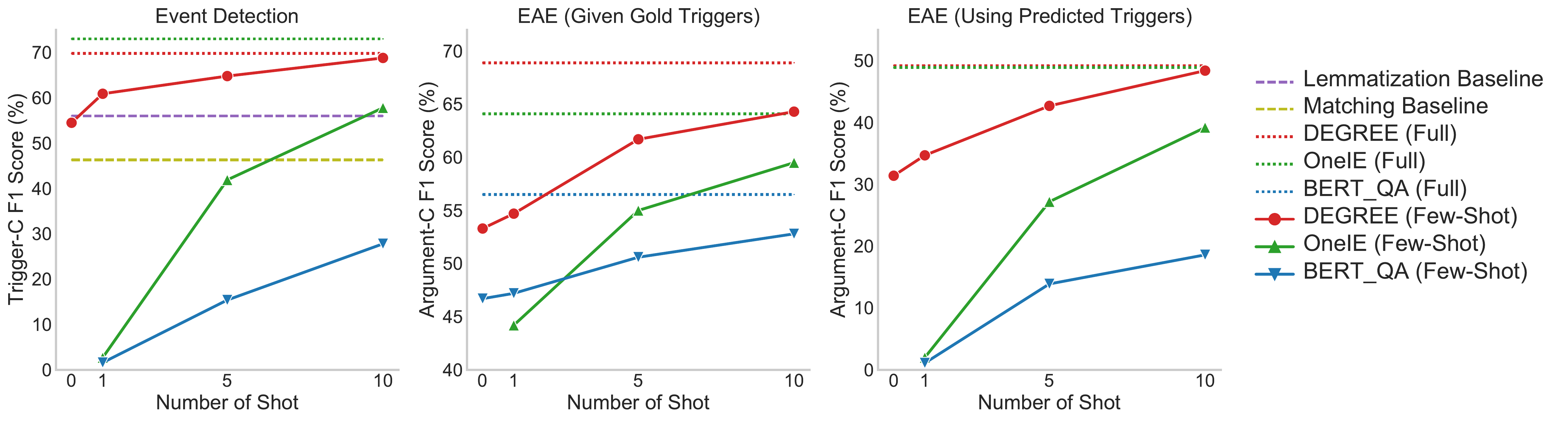}
    \caption{Results for top common 10 event types.}
    \end{subfigure}
    \caption{The zero/few-shot experimental results. \textit{\textbf{Left}}: The result for the models on event detection task with the scores reported in trigger classification F1.
    \textit{\textbf{Middle}}: The models are tested under the scenario of given gold trigger and evaluated with argument classification criterion.  \textit{\textbf{Right}}: The results for the models to perform event extraction task, which aims to predict triggers and their corresponding arguments (we report the argument classification F1).}
    \label{fig:zero_few_shot}
\end{figure*}

\paragraph{Experimental results.}
Figure~\ref{fig:zero_few_shot}, Table \ref{tab:zeroee}, and Table \ref{tab:zeroeae} show the results of $n=5$ and $n=10$. 
From the two subfigures in the left column, we see that \edmodel{} achieves promising results in the zero-shot setting. In fact, it performs better than BERT\_QA trained in the 10-shot setting and OneIE trained in the 5-shot setting. This demonstrates the great potential of \edmodel{} to discover new event types.
Interestingly, we observe that our two proposed baselines perform surprisingly well, suggesting that the trigger annotations in ACE05-E are actually not diverse. 
Despite their impressive performance, \edmodel{} still outperforms the matching baseline by over 4.7\% absolute trigger classification F1 in both $n=5$ and $n=10$ cases in zero-shot scenario.
Additionally, with only one training instance for each unseen type, \edmodel{} can outperform both proposed baselines.

Next, we compare the results for the event argument extraction task.
From the two middle subfigures, we observe that when given gold triggers, our model performs much better than all baselines with a large margin.
Lastly, we train models for both trigger and argument extraction and report the final argument classification scores in the two right subfigures. 
We justify that our model has strong generalizability to unseen event types and it can outperform BERT\_QA and OneIE even when they are both trained in 5-shot settings.


\clearpage

\begin{table*}[t!]
\centering
\small
\setlength{\tabcolsep}{5pt}
\begin{tabular}{|ll|cc|cc|cc|cc|}
    \hline
    \multicolumn{10}{|c|}{Event Extraction} \\
    \hline
    \multirow{2}{*}{Trigger} & \multirow{2}{*}{Argument} & \multicolumn{4}{c|}{Common 5} & \multicolumn{4}{c|}{Common 10} \\
    \cline{3-10}
    & &  Tri-I &  Tri-C & Arg-I &  Arg-C &  Tri-I &  Tri-C & Arg-I &  Arg-C \\
    \hline
    \multicolumn{2}{|l|}{Matching Baseline}        & 42.7 & 42.1 & - & - & 46.3 & 46.3 & - & - \\
    \multicolumn{2}{|l|}{Lemmatization Baseline}  & 51.5 & 50.2 & - & - & 56.6 & 56.0 & - & - \\
    \hline
    \multicolumn{2}{|l|}{BERT\_QA 1-shot}   & 10.0 & 1.4 & 1.3 & 1.3 &  8.2 &  1.6 &   1.1 &  1.1 \\
    \multicolumn{2}{|l|}{BERT\_QA 5-shot}   & 14.0 & 12.6 & 11.1 & 10.8 &  20.8 & 15.4 & 14.6 & 13.9 \\
    \multicolumn{2}{|l|}{BERT\_QA 10-shot} & 37.8 & 33.5 & 22.9 & 22.1 &  32.0 & 27.8 & 19.5 & 18.6 \\
    \multicolumn{2}{|l|}{OneIE 1-shot}         & 4.2 & 4.2 & 1.5 & 1.5 & 4.1 & 2.7 & 2.0 & 2.0 \\
    \multicolumn{2}{|l|}{OneIE 5-shot}         & 39.3 & 38.5 & 24.8 & 22.8 & 41.9 & 41.9 & 29.7 & 27.2 \\
    \multicolumn{2}{|l|}{OneIE 10-shot}         & 54.8 & 53.3 & 36.0 & 34.9 & 61.5 & 57.8 & 41.4 & 39.2  \\
    \edmodel{} 0-shot        & \eaemodel{} 0-shot        & 53.3 & 46.8 & 29.6 & 25.1 & 60.9 & 54.5 & 42.0 & 31.4 \\
    \edmodel{} 1-shot   & \eaemodel{} 1-shot            & 60.1 & 53.3 & 38.8 & 31.6 & 61.2 & 60.9 & 41.1 & 34.7 \\
    \edmodel{} 5-shot        & \eaemodel{} 5-shot        & 57.8 & 55.5 & 40.6 & 36.1 & 65.8 & 64.8 & 45.3 & 42.7 \\
    \edmodel{} 10-shot        & \eaemodel{} 10-shot        & 63.8 & 61.2 & 46.0 & 42.0 & 72.1 & 68.8 & 52.5 & 48.4 \\
    \hline
    \multicolumn{2}{|l|}{OneIE (Full)}         & 72.7 & 70.5 & 52.3 & 49.9 & 74.5 & 73.0 & 51.2 & 48.9  \\
    \edmodel{} (Full)        & \eaemodel{} (Full)        & 68.4  & 66.0 & 51.9 & 48.7 & 72.0 & 69.8 & 52.5 & 49.2 \\
    \hline
 
\end{tabular}
\caption{Full results of zero/few-shot event extraction on ACE05-E.}
\label{tab:zeroee}
\end{table*}

\begin{table*}[t!]
\centering
\small
\setlength{\tabcolsep}{5pt}
\begin{tabular}{|ll|cc|cc|cc|cc|}
    \hline
    \multicolumn{10}{|c|}{Event Argument Extraction} \\
    \hline
    \multirow{2}{*}{Trigger} & \multirow{2}{*}{Argument} & \multicolumn{4}{c|}{Common 5} & \multicolumn{4}{c|}{Common 10} \\
    \cline{3-10}
    & &  Tri-I &  Tri-C & Arg-I &  Arg-C &  Tri-I &  Tri-C & Arg-I &  Arg-C \\
    \hline
    Gold Triggers   & BERT\_QA 0-shot   & 100.0& 100.0& 55.8 & 37.9 & 100.0& 100.0& 57.2& 46.7 \\
    Gold Triggers   & BERT\_QA 1-shot   & 100.0& 100.0& 55.8 & 44.3 & 100.0& 100.0& 57.8 & 47.2 \\
    Gold Triggers   & BERT\_QA 5-shot   & 100.0& 100.0& 56.6 & 49.6 & 100.0& 100.0& 59.1 & 50.6 \\
    Gold Triggers   & BERT\_QA 10-shot   & 100.0& 100.0& 58.8 & 52.9 & 100.0& 100.0& 60.5 & 52.8 \\
    Gold Triggers   & OneIE 1-shot   & 100.0& 100.0& 40.9 & 36.5 & 100.0& 100.0 & 48.3 & 44.2\\
    Gold Triggers   & OneIE 5-shot   & 100.0& 100.0& 55.6 & 51.4 & 100.0& 100.0 & 58.6 & 55.0\\
    Gold Triggers   & OneIE 10-shot   & 100.0& 100.0& 59.4 & 56.7 & 100.0& 100.0 & 62.0 & 59.5 \\    
    Gold Triggers    & \eaemodel{} 0-shot        & 100.0& 100.0& 56.1 & 48.0 & 100.0& 100.0& 66.5 & 53.3 \\
    Gold Triggers    & \eaemodel{} 1-shot        & 100.0& 100.0& 65.2 & 55.2& 100.0& 100.0& 65.4& 54.7 \\
    Gold Triggers    & \eaemodel{} 5-shot        & 100.0& 100.0& 70.9 & 62.2 & 100.0& 100.0& 68.0& 61.7  \\
    Gold Triggers    & \eaemodel{} 10-shot        & 100.0& 100.0& 71.1 & 64.2 & 100.0& 100.0& 71.6 & 64.3 \\
    \hline
    Gold Triggers   & BERT\_QA (Full)   & 100.0& 100.0& 63.1 & 57.9 & 100.0& 100.0& 62.1& 56.5 \\
    Gold Triggers   & OneIE (Full)   & 100.0& 100.0& 70.8 & 66.4 & 100.0& 100.0& 67.9& 64.1 \\
    Gold Triggers    & \eaemodel{} (Full)        & 100.0& 100.0& 74.5 & 70.6 & 100.0& 100.0& 73.6 & 68.9 \\
    \hline
 
\end{tabular}
\caption{Full results of zero/few-shot event argument extraction on ACE05-E.}
\label{tab:zeroeae}
\end{table*}

\clearpage

\end{document}